\documentclass[journal]{IEEEtran}

\usepackage{cite}
\usepackage{amsmath,amssymb,amsfonts}
\usepackage{bm}
\usepackage{graphicx}
\usepackage{multirow} 
\usepackage{tabularx}
\usepackage{dblfloatfix}
\usepackage{amsmath}
\usepackage{booktabs}
\usepackage{algorithmic}
\usepackage[hidelinks]{hyperref}
\usepackage{orcidlink}

\makeatletter
\patchcmd{\@makecaption}
  {\ifdim \xfigwd >\columnwidth}
  {\ifdim \xfigwd >\linewidth}
  {}{}
\patchcmd{\@makecaption}
  {\mbox{}\hfill\begin{tabular}{@{}l@{}}\noindent\raggedright\unhbox\@tempboxa\figcapfont#2\end{tabular}\hfill\mbox{}}
  {\noindent\makebox[\linewidth][c]{%
     \parbox[t]{\linewidth}{\raggedright\unhbox\@tempboxa\figcapfont#2}%
   }}
  {}{}
\makeatother

\def\BibTeX{{\rm B\kern-.05em{\sc i\kern-.025em b}\kern-.08em
    T\kern-.1667em\lower.7ex\hbox{E}\kern-.125emX}}

\newcommand{\sg}{\operatorname{stopgrad}}

\newcommand{\Agg}{\operatorname{Agg}}

\begin{document}

\title{Causally-Constrained Probabilistic Forecasting for Time-Series Anomaly Detection}

\author{
  Pooyan Khosravinia\,\orcidlink{0000-0002-4968-5800},
  João Gama\,\orcidlink{0000-0003-3357-1195},
  Bruno Veloso\,\orcidlink{0000-0001-7980-0972}%
\thanks{}%
}

\maketitle

\begin{abstract}
Anomaly detection in multivariate time series is important for monitoring complex industrial systems, where failures often appear through delayed interactions among multiple sensors. Although recent deep learning methods such as graph neural networks and Transformers achieve strong detection accuracy, many of them mainly learn correlations and provide limited insight into why an anomaly occurs, or which variables are most responsible. To address this limitation, this study proposes the Causally Guided Transformer (CGT), a causal graph-guided probabilistic forecasting framework for multivariate time-series anomaly detection. CGT first uses a time-lagged causal graph as a structural prior to identify which past variables are most relevant for forecasting each target variable. Each target-specific forecasting block then predicts a probability distribution for the next value which allows anomalies to be detected when observations are unlikely under the predicted distribution. To improve robustness, CGT also includes a gradient-isolated residual auxiliary path that can use additional correlational information, while a safety-gated mechanism limits its contribution when reliability is low. Root-cause variables are ranked using per-dimension probabilistic attribution. Experiments on SMD, ASD, and SWaT show that CGT achieves F1-scores of 95.32\%, 96.19\%, and 83.91\%, respectively, while obtaining the best AUROC on ASD and SWaT. These results show that causal structural priors can improve both detection robustness and interpretability in multivariate sensor systems.
\end{abstract}


\section{Introduction}
\label{sec:introduction}
Industrial systems are inherently complex and are usually monitored through a large number of sensors in domains such as industrial IoT, manufacturing, aerospace, power grids, and healthcare. These sensors produce multivariate time-series data that capture the evolving behavior of a system. Detecting normal and anomalous patterns in such data is essential for early fault detection, improved safety, and predictive maintenance, which helps reduce costly downtime and extend equipment lifespan \cite{chandola2009anomaly}. However, anomaly detection in multivariate sensor data remains challenging because of high dimensionality, noise, non-stationarity, and strong spatial and temporal dependencies among variables \cite{gupta2013outlier}.

Existing statistical, machine learning, and deep learning methods have achieved strong performance in many anomaly detection settings. In particular, recurrent neural networks (RNNs), graph-based models, and Transformers can capture complex temporal behavior and cross-sensor dependencies. However, most of these methods rely primarily on correlation patterns rather than an explicit cause-and-effect structure. Strong correlation does not necessarily imply that one variable causes or explains another, and this can lead to misleading relationships, false alarms, and limited insight into the true source of an anomaly \cite{su2019robust,runge2019inferring}. As a result, many current models can detect abnormal behavior but struggle to determine why it occurred and which component is most likely responsible. 

Therefore, recent work has begun to incorporate causal structure into anomaly detection pipelines. Causal relationships among sensors can be derived from domain knowledge or learned from data, and provide a more interpretable basis for evaluating whether an observation is consistent with expected system behavior \cite{sun2025unifying}. For example, if sensor A causally influences sensor B, then an extreme change in B while A remains normal may indicate an anomaly in B or in the mechanism linking A to B. By contrast, a simultaneous change in both A and B may be expected if the change in B is a downstream effect of A. Causal modeling can therefore improve anomaly detection by focusing on stable mechanisms and can support root-cause analysis by distinguishing primary anomalies from propagated effects.

This study proposes a Causally Guided Transformer (CGT) model to provide multivariate time-series anomaly detection in which a deep sequence model is informed by an explicit causal graph among variables. Rather than treating the sensor network as an unstructured collection of signals, the approach uses causal structure to guide attention, prediction, and interpretation. The goal is not to replace deep representation learning with a fixed graph, but to use causal knowledge as an inductive bias that encourages more interpretable and causally consistent learning. By detecting behaviors that violate expected cause-effect relationships and tracing anomaly propagation along causal links, such a model can better distinguish root causes from secondary effects. To make the proposed idea intuitive, CGT can be viewed as a graph-guided forecasting model. Instead of allowing every sensor to influence every other sensor freely, the model first uses a time-lagged causal graph to identify which past variables are most relevant for predicting each target variable. This graph acts as a structural prior for forecasting. When the actual observation is unlikely under the predicted distribution, the model assigns a high anomaly score. The same forecasting errors and input-clamping analysis are then used to rank variables that most strongly contribute to the detected anomaly. In the experiments, we evaluate CGT on three widely used multivariate anomaly-detection benchmarks: the Server Machine Dataset (SMD), the Application Server Dataset (ASD), and the Secure Water Treatment (SWaT) dataset. 

The main contributions of this study are summarized as follows:

\begin{itemize}
\item We introduce CGT, a causally-guided probabilistic forecasting method for multivariate time-series anomaly detection that integrates a time-lagged causal graph prior with Transformer-based sequence modeling.

\item We propose a safety aware auxiliary learning mechanism that captures useful residual dependencies beyond the causal graph through a stop-gradient auxiliary path and reliability gated score blending.

\item We formulate anomaly detection and root-cause analysis within a unified probabilistic framework, using Gaussian predictive likelihoods for detection and both standardized likelihood attribution and counterfactual-style clamping for diagnosis.

\item We validate the proposed method on ASD, SMD and SWaT through benchmark comparison and ablation analysis, showing strong detection performance and improved variable-level attribution quality.
\end{itemize}

The rest of this paper is organized as follows. We first introduce the background on multivariate time-series anomaly detection and causal discovery, and then review the related work. Next, we describe the proposed methodology, followed by the experimental setup and the obtained results.
The implementation presented in this work is available at \url{https://github.com/p-khn/CGT-V1}

\section{Background}
\label{sec:background}
This section introduces the main concepts needed to understand the proposed CGT framework, including multivariate time-series anomaly detection, probabilistic forecasting, time-lagged causal graphs, and causally guided attribution.
\subsection{Multivariate Sensor Time Series}

A multivariate sensor time series is represented as
\begin{equation}
\mathbf{X} \in \mathbb{R}^{T \times D},
\end{equation}
where $T$ is the sequence length and $D$ is the number of variables. The system state at time $t$ is
\begin{equation}
\mathbf{x}_t = \big[x_t^{(1)}, x_t^{(2)}, \dots, x_t^{(D)}\big]^\top,
\end{equation}
where $x_t^{(i)}$ denotes the reading of sensor $i$ at time $t$.

Multivariate sensor time series exhibit three characteristics that make anomaly detection difficult.

\textit{Dependencies among variables:} Sensors may be statistically dependent because of shared operating conditions, physical coupling, or direct influence. In many systems, some dependencies are merely correlational, whereas others reflect true causal mechanisms. These inter-sensor relationships allow faults or disturbances to propagate across variables, making joint modeling essential \cite{deng2021graph,zhang2022grelen}.

\textit{Temporal dynamics:} Sensor readings evolve over time and often exhibit trends, seasonality, autocorrelation, and regime changes. Real-world systems are rarely stationary, and sensor streams may also contain missing values or noise, all of which complicate modeling \cite{gupta2013outlier}.

\textit{High dimensionality:} Modern sensor networks may contain dozens or hundreds of variables. As the number of sensors grows, modeling cross-variable interactions becomes increasingly difficult.

This study focuses on unsupervised anomaly detection in multivariate time series, where anomalies are rare and labeled fault examples are limited, while normal operating data are relatively abundant.

\subsection{Time-Series Anomalies and Detection Principles}

Anomalies in time-series data are observations or patterns that deviate from expected normal behavior \cite{chandola2009anomaly,gupta2013outlier}. In practice, they are often categorized into three broad types.

Time-series anomalies are commonly categorized as point, contextual, or collective anomalies. Point anomalies correspond to individual observations that deviate strongly from normal behavior, contextual anomalies are abnormal only under a specific temporal or operational context, and collective anomalies correspond to abnormal patterns that emerge over a sequence of observations.

Time-series anomaly detection methods are commonly organized by learning setting and modeling principle. In supervised settings, models are trained using labeled anomalies. However, because anomalies are rare and highly diverse, unsupervised and semi-supervised approaches are more common. Semi-supervised methods typically train on normal data only and treat anomaly detection as novelty detection or one-class classification.

Reconstruction-based methods learn to reproduce normal patterns and flag observations with high reconstruction error as anomalous. Autoencoders (AEs) and variational autoencoders (VAEs) are common examples. Forecast-based methods instead predict future values from past observations and treat large prediction residuals as anomaly scores. Classical models such as ARIMA and Kalman filters, as well as modern RNN-based predictors, fall into this category. Probabilistic methods assign a likelihood to each observation or sequence and mark low-probability events as anomalous; examples include Hidden Markov Models, Gaussian processes, and deep generative models.

Other approaches include distance-based and clustering-based methods, such as nearest-neighbor scoring and isolation forests. These methods can work well in lower-dimensional settings but often struggle with long, high-dimensional multivariate sequences because defining meaningful similarity becomes difficult. Across all of these families, one central challenge remains: anomaly detection must account not only for temporal structure within each sensor but also for dependencies among sensors. This motivates the use of structural information, including causal relationships, to guide detection.

\subsection{Causal Inference and Causal Discovery in Time Series}

Causal inference seeks to explain how variables influence one another rather than simply how they co-vary. These relationships are commonly represented by a directed acyclic graph (DAG), in which nodes correspond to variables and directed edges indicate direct causal effects. In time-series settings, causal structure is temporal: a variable at an earlier time may influence another variable at a later time. For example, $X^{(i)}_{t-\ell}$ may causally influence $X^{(j)}_t$.

A structural causal model (SCM) describes each variable as a function of its direct causes and exogenous noise:
\begin{equation}
X^{(j)}_t = f^{(j)}(Pa^{(j)}_t, U^{(j)}_t),
\end{equation}
where $Pa^{(j)}_t$ denotes the causal parents of $X^{(j)}_t$ and $U^{(j)}_t$ is an exogenous noise term. A key advantage of SCMs is that they support reasoning about interventions. Rather than asking what typically co-occurs with a variable, one can ask what happens when a variable is actively changed. In industrial systems, this perspective is valuable for understanding fault propagation and the effects of intervention.

A foundational concept in temporal causality is Granger causality \cite{granger1969investigating}. A time series $X$ is said to Granger-cause another series $Y$ if past values of $X$ improve the prediction of future values of $Y$ beyond what can be achieved using the past of $Y$ and other relevant variables alone \cite{runge2019inferring}. Classical Granger tests rely on vector autoregressive (VAR) models, whereas more recent methods extend the idea to nonlinear settings through neural networks, sparsity penalties, or information-theoretic criteria. Neural Granger Causality, for example, uses sparse neural networks to identify predictive links while suppressing irrelevant ones \cite{tank2022neural}.

Beyond pairwise Granger analysis, a broader literature addresses causal discovery in multivariate time series. Constraint-based methods, such as PC and its time-series variants, use conditional independence tests to infer graph structure under assumptions of causal sufficiency. PCMCI, which stands for Peter and Clark Momentary Conditional Independence, is a constraint-based causal discovery algorithm for multivariate time series. It first identifies candidate lagged parents for each variable and then tests momentary conditional independencies to remove indirect or statistically unsupported links. The output is a time-lagged directed graph whose edges indicate candidate temporal causal dependencies, such as \((j,\ell)\rightarrow i\), meaning that variable \(j\) at lag \(\ell\) is selected as a candidate parent of target variable \(i\). PCMCI extends constraint-based causal discovery to time-lagged and potentially nonlinear settings while improving statistical efficiency in higher-dimensional problems \cite{runge2019inferring}. Score-based methods instead search for the graph that best explains the data according to a chosen objective. NOTEARS formulates static DAG learning as a continuous optimization problem with an acyclicity constraint, while DYNOTEARS extends this idea to time-series data by jointly estimating contemporaneous and time-lagged dependencies in a dynamic Bayesian network \cite{pamfil2020dynotears}. Functional causal models such as LiNGAM impose stronger assumptions about the data-generating process, for example, linearity and non-Gaussian noise.

Each class of methods has limitations. Constraint-based methods may become unreliable in noisy or high-dimensional settings. Score-based methods can be computationally expensive, and functional causal models may impose assumptions that are too restrictive for complex industrial systems \cite{xin2023causalrca}. In addition, many real systems contain latent confounders or partially observed variables, so the learned graph should often be treated as a useful but imperfect structural prior rather than absolute ground truth.

Causal discovery is highly relevant to anomaly detection because it helps separate direct effects from indirect effects. In a system where sensor S1 influences both S2 and S3, all three variables may be correlated, but a causal model aims to determine whether S1 is the common driver and whether the apparent dependence between S2 and S3 is only indirect. This distinction matters for diagnosis: changing S1 may affect S2 and S3, while intervening on S2 should not affect S1. Such reasoning is not available in purely correlational models.

\subsection{Causally Guided Anomaly Detection}

Causally guided anomaly detection uses causal knowledge, typically represented as a causal graph, to improve both anomaly detection and interpretation in multivariate time-series data. Instead of assessing each observation only by how unusual it appears in isolation or by correlation with other variables, the detector evaluates whether it is consistent with the expected behavior of its causal parents.

A key advantage of causal modeling is that it allows anomaly scores to be conditioned on the variables that directly influence each sensor. Instead of evaluating a variable in isolation, the model asks whether its value is reasonable given the state of its causal parents. For example, if sensor (B) changes unexpectedly while its parent (A) remains within its normal range, this may indicate a fault in (B) or a disruption in the mechanism linking (A) to (B). However, if both (A) and (B) change in a way that is consistent with the learned causal relationship, then the behavior of (B) may be normal, even if it appears unusual under a correlation-based model.

Causal models can also detect structural anomalies, which occur when the expected relationship between variables is violated. Such anomalies are different from isolated measurement errors. A faulty sensor may produce an abnormal reading without affecting the rest of the system, whereas a genuine system fault may propagate through several connected variables according to the underlying dependency structure \cite{sun2025unifying}. Recognizing this difference helps distinguish local sensor errors from broader system-level disturbances.

A further benefit is improved interpretability and root-cause identification. Because causal graphs represent directional relationships, they provide a natural way to explain how an anomaly spreads through a system. When several sensors become abnormal, the graph can be used to trace the disturbance toward its most plausible upstream source. As a result, a causally guided detector can do more than flag abnormal variables: it can distinguish likely primary faults from the downstream effects they produce.

In practice, causally guided anomaly detection begins by learning or specifying a causal graph from historical data, then building a detector informed by this structure. One approach is to define a probabilistic model over the graph and measure how well new observations fit the expected dependencies \cite{deng2021graph}. Another is to use parent-child relationships to generate predictions, treating large mismatches between expected and observed behavior as anomaly signals. The causal graph can also be embedded into deep learning architectures, such as autoencoders or Transformers, so that sequence learning is shaped by structured information about variable influence.

Although the learned graph may be imperfect, even an approximate causal prior can provide useful inductive bias. It can reduce reliance on spurious correlations, improve robustness under changing conditions, and produce more actionable explanations. For example, in a water treatment system, a failed valve may cause downstream changes in flow rate and tank level. A causally guided detector can interpret this pattern as a single fault with propagated effects, whereas a purely black-box detector may simply mark all affected sensors as anomalous without explanation.

\section{Related Work}
\label{sec:related}
This section reviews existing methods relevant to multivariate timeseries anomaly detection, including reconstruction-based, graph-based, probabilistic, Transformer-based, and causality-guided approaches.
\subsection{Anomaly Detection in Multivariate Time Series}

Classical approaches to timeseries anomaly detection often rely on statistical models that capture temporal structure for each variable individually or after dimensionality reduction. Forecasting-based methods, such as ARIMA, identify anomalies through large prediction residuals. For multivariate data, vector autoregressive models and residual-based monitoring can capture linear interdependencies among variables. Subspace methods such as Principal Component Analysis (PCA) learn low-dimensional structure from normal data and identify anomalies through reconstruction error or projection onto low-variance directions. For example, PCA has been used successfully to detect unusual traffic patterns in network monitoring \cite{lakhina2004diagnosing}. These methods offer some interpretability, but they are limited in their ability to model nonlinear dependencies and may be sensitive to distribution shift.

Deep learning has substantially expanded the modeling capacity of anomaly detection systems for multivariate time series. Recurrent neural networks, including LSTMs and GRUs, are widely used for modeling temporal dependencies. LSTM encoder-decoder architectures detect anomalies by reconstructing normal windows and flagging large reconstruction errors \cite{malhotra2016lstm}. LSTM forecasting models have also been used for anomaly detection in aerospace telemetry through dynamic thresholding on prediction residuals \cite{hundman2018}. Frequency-domain models have been investigated for efficient time-series anomaly detection. FITS employs complex frequency-domain interpolation to reconstruct time-series segments and identifies anomalies based on reconstruction error, while utilizing a minimal number of parameters \cite{xu2024fits}. However, FITS remains a reconstruction-based representation model and does not explicitly capture causal dependencies or facilitate root-cause attribution. Recent frequency-domain approaches further improve multivariate time-series anomaly detection by modeling abnormal subsequences in spectral space. CATCH patchifies the frequency domain into fine-grained bands and uses a channel fusion module to capture adaptive channel correlations, achieving strong performance on diverse point and subsequence anomalies \cite{wu2025catch}. However, it remains a reconstruction-based representation model and does not explicitly model causal dependencies or root-cause attribution.

Several deep models explicitly represent multivariate time series as structured objects. MSCRED converts sensor relationships into signature matrices and applies a CNN-LSTM architecture to reconstruct them, thereby detecting abnormal correlation patterns \cite{zhang2019mscred}. Graph-based methods, such as GDN, represent sensor relationships as a graph and compute anomaly scores based on deviations from these learned dependencies \cite{deng2021graph}. MTAD-GAT uses graph attention mechanisms over a fully connected sensor graph to capture variable interactions while combining temporal modeling with attention-based weighting \cite{zhao2020mtadgat}. These models highlight the importance of inter-sensor structure, but the learned relationships are still typically correlational rather than explicitly causal.

Generative methods form another important family. GAN-based models such as MAD-GAN learn the distribution of normal subsequences and use discriminator-based signals to identify anomalies. Variational models such as InterFusion similarly model latent normal behavior and use deviations from learned distributions as anomaly scores \cite{li2021interfusion}. While such models can capture complex data distributions, they often require careful calibration and may be difficult to interpret.

More recently, Transformer-based approaches have become prominent because of their ability to capture long-range dependencies. The Anomaly Transformer, for example, uses self-attention to learn associations among time points and has achieved strong results on benchmark datasets \cite{xu2021anomaly}. Self-supervised contrastive methods have also been proposed for time-series anomaly detection. DCdetector uses a dual-attention contrastive framework to compare patch-wise and in-patch representations, detecting anomalies through representation discrepancy rather than reconstruction error \cite{yang2023dcdetector}. However, it still focuses on representation-level abnormality and does not explicitly model causal parent-child dependencies or causal root-cause attribution. Despite their strong empirical performance, most of these models remain fundamentally correlational. They can identify unusual patterns, but they do not explicitly distinguish direct causes from indirect effects. Recent advancements in deep time-series backbone architectures have enhanced anomaly detection by enabling the learning of more comprehensive temporal representations. For instance, TimesNet transforms one-dimensional time series into two-dimensional tensors based on dominant periods and applies two-dimensional kernels to capture both intra-period and inter-period variations \cite{wu2023timesnet}. Nevertheless, TimesNet primarily emphasizes temporal representation learning rather than explicit modeling of causal dependencies or root-cause attribution. By comparison, CGT incorporates a time-lagged causal graph prior to inform probabilistic forecasting and diagnostic processes.

\subsection{Causal Inference for Multivariate Time Series}

Causal discovery in time series has been widely studied in fields such as econometrics, neuroscience, climate science, and system monitoring. Granger causality remains a core starting point, but modern methods extend it to high-dimensional and nonlinear systems. Neural Granger Causality uses sparse neural architectures to infer predictive directional relationships \cite{tank2022neural}. Transfer entropy provides an information-theoretic notion of directional influence and has been applied to complex temporal systems, although it can be computationally demanding in large networks \cite{schreiber2000measuring}. Sparse VAR-based approaches such as Lasso-Granger attempt to scale causal discovery by combining autoregressive modeling with variable selection \cite{arnold2007temporal}.

Constraint-based approaches such as PC and FCI have also been extended to temporal settings by treating lagged variables as distinct nodes. PCMCI is a particularly important example because it combines conditional independence testing with a time-series-specific design that improves feasibility in large systems and supports nonlinear dependence testing \cite{runge2019inferring}. Score-based approaches such as GES, NOTEARS, and temporal extensions such as Dynotears instead optimize a graph-scoring objective subject to structural constraints \cite{zheng2018dags}. Functional causal models, including LiNGAM, assume specific noise and functional forms in order to identify directionality. Recent deep learning approaches have also been developed for temporal causal discovery. CausalFormer uses a causality-aware Transformer and a decomposition-based causality detector to infer temporal causal relations and time lags from trained predictive models \cite{kong2024causalformer}. This line of work shows the growing use of interpretable neural models for learning time-lagged causal structure, whereas CGT learns such structure as a prior for probabilistic anomaly detection and attribution.

Although these methods have significantly advanced time-series causal discovery, their practical use in industrial anomaly detection remains challenging. Noise, nonlinearity, limited samples, and partial observability can all degrade graph quality. Accordingly, learned causal graphs in such settings should often be interpreted as informative structural priors rather than exact representations of the true system. 

\subsection{Causality-Guided Anomaly and Fault Analysis}

Recent work at the intersection of anomaly detection and causal analysis has shown that structural reasoning can improve root-cause localization and anomaly interpretation, especially in complex operational systems. In microservices and IT operations, methods such as CauseInfer and MicroDiag infer dependency structures among system metrics and use them to trace anomaly propagation paths \cite{chen2014causeinfer,wu2021microdiag}. CausalRCA extends this idea by learning more expressive causal graphs and using them to identify likely root causes after anomalies are detected, improving localization accuracy on benchmark data \cite{xin2023causalrca}. More recently, AERCA integrates Granger causal discovery with root-cause analysis by treating anomalies as interventions on exogenous variables in multivariate time series. It learns normal exogenous-variable distributions through an encoder-decoder framework and identifies root causes by detecting which inferred exogenous variables deviate from normal behavior \cite{han2025aerca}. However, AERCA focuses mainly on causal discovery and root-cause localization, whereas CGT integrates causal graph priors directly into probabilistic forecasting-based anomaly detection and attribution. GCAD detects multivariate time-series anomalies by monitoring deviations in dynamic Granger causality patterns learned from gradients of a deep predictor \cite{liu2025gcad}. Unlike reconstruction- or prediction-error methods, it treats changes in causal dependency structure as the anomaly signal. However, GCAD focuses on anomaly detection through causal-pattern deviation, whereas CGT uses a time-lagged causal graph prior for probabilistic forecasting and variable-level attribution.

Related work has also explored structure-aware anomaly scoring. GReLeN, for example, learns a probabilistic relational graph among variables and uses it to evaluate how well new observations align with expected dependency patterns \cite{zhang2022grelen}. Results on datasets such as SWaT and WADI suggest that modeling inter-variable structure can improve both detection and diagnosis.

A key limitation of much of this literature is that anomaly detection and causal analysis are often treated as separate stages: a model first detects anomalies, and only afterwards is causal reasoning applied to explain them. In addition, many graph-based detectors rely on structural dependencies that are useful in practice but do not necessarily represent true causal mechanisms. Even so, this line of work supports an important conclusion: incorporating structured dependencies can improve anomaly detection and interpretation relative to unstructured approaches. This leaves an open opportunity to integrate causal structure more directly into deep anomaly detectors themselves, rather than using it only after detection or as a loosely connected prior. In contrast to these methods, CGT integrates a time-lagged causal graph prior directly into a probabilistic Transformer forecasting architecture, while also providing likelihood-based anomaly scoring and graph-guided root-cause attribution.

\section{Methodology}
\label{sec:method}
\begin{figure*}[t!]
    \centering
    \includegraphics[width=\textwidth]{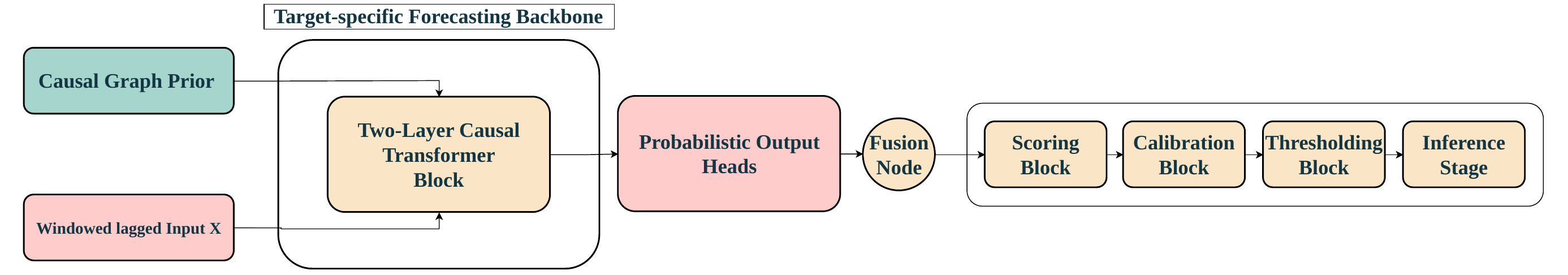}
    \caption{Overall pipeline of the proposed causal-probabilistic anomaly detection framework. A causal graph prior and windowed lagged multivariate input are processed by target-specific forecasting blocks to produce probabilistic predictions. The predictions are then converted into anomaly scores, calibrated through safe auxiliary activation, and thresholded for anomaly detection.}
    \label{fig:overview}
\end{figure*}

This study proposes a probabilistic approach to detecting anomalies in time-series data that preserves causal structure. The architecture contains forecasting blocks. It learns each target variable using a separate forecasting block, which contains an encoder, a latent-variable component, and a prediction head. Each block includes two distinct computational paths, the causal path and the auxiliary path. The first path updates the encoder, the latent module, and the primary head; its input is the parents' features. The auxiliary path uses additional correlations from other variables if needed, even if those variables are not direct causes. However, these extra path contributions are activated only if a safety test indicates minimal reliance on non-parent features, so their learning does not affect the main causal learning process. Otherwise, it falls back to the purely causal anomaly scoring. Stop-gradient isolation is a standard mechanism to prevent unintended interference between branches during multi-objective training. It has been widely used in related the multi-branch representation learning settings \cite{grill2020bootstrap,chen2021simsiam}. Figure~\ref{fig:overview} shows the overall workflow of the proposed method.


\subsection{Problem Formulation}
\label{sec:problem}

Consider a multivariate time series $\{x_t\}_{t=1}^{T}$, where $x_t \in \mathbb{R}^{D}$ represents measurements collected for $D$ sensors. For each target variable \(i \in \{1,\dots,D\}\), the model estimates the one-step-ahead conditional density \(p(x_t^{(i)} \mid \mathcal{H}_t)\), where \(\mathcal{H}_t\) denotes the windowed lagged history constructed from past observations across all sensors up to maximum lag \(\tau_{\max}\). The predictive distribution is parameterized as a Gaussian:
\begin{equation}
p\!\left(x_t^{(i)} \mid \mathcal{H}_t\right)
=
\mathcal{N}\!\left(x_t^{(i)};\mu_{t,i},\sigma^2_{t,i}\right),
\end{equation}
where \(\mu_{t,i}\) and \(\sigma^2_{t,i}\) are produced by the target-specific forecasting block.
Anomalies are detected using the negative log-likelihood (NLL). So, observations that yield high NLL under the learned predictive distribution are flagged as anomalous.
Throughout the methodology section, we denote the prediction target for variable \(i\) at time \(t\) as \(y_{t,i}=x_t^{(i)}\). Quantities with superscript \((i)\) are produced by the target-specific forecasting block assigned to variable \(i\). The causal predictive density produced by the graph-constrained branch is denoted by \(p_c^{(i)}\), while the residual auxiliary predictive density is denoted by \(p_o^{(i)}\). Throughout the methodology section, \(X_t\) denotes the windowed lagged input matrix, while \(x_t^{(i)}\) denotes the scalar observation of variable \(i\) at time \(t\). We denote the prediction target as \(y_{t,i}=x_t^{(i)}\).
\subsection{Windowed Lagged Input Construction}

\label{sec:window}
A windowed lag design matrix is constructed to align the forecasting model with the structure of the time-lagged causal graph.
Here, $W$ denote the window length and $\tau_{\max}$ represents the maximum lag.
The set of sensor--lag pairs is defined as follows. 
\begin{equation}
\mathcal{P}=\{(j,\ell): j \in \{1,\dots,D\},\ \ell \in \{1,\dots,\tau_{\max}\}\},
\end{equation}
with $P = |\mathcal{P}| = D\tau_{\max}$.
At each time $t$, the model input consists of a matrix
\begin{equation}
X_t \in \mathbb{R}^{W \times P},
\end{equation}
The column corresponding to the sensor--lag pair $(j,\ell)$ is formed by a length-$W$ history shifted by $\ell$ the steps:
\begin{equation}
X_t[:,(j,\ell)] = \big[x_{t-W-\ell}^{j},\ x_{t-W-\ell+1}^{j},\ \dots,\ x_{t-1-\ell}^{j}\big]^\top.
\end{equation}
Each of the $W$ rows represents a token that contains all the sensors across all lags at the same relative position within the window.
All channels are scaled using a Min--Max transformation, which is fitted on the training data and subsequently applied to test sequences. Because this construction requires $W$ historical points for each lag and valid inputs exist only for $t \ge W+\tau_{\max}$.

\subsection{Causal Graph Prior and Parent Mask}
\label{sec:pi}
A Time-lagged causal discovery method, based on the PCMCI algorithm, is used to estimate directed, lagged dependencies of the form $(j,\ell)\rightarrow i$. This notation indicates that the sensor $j$ at lag $\ell$ is a candidate cause of target variable $i$ \cite{runge2020discovering}.
For each target variable $i$, a binary parent mask is constructed.
\begin{equation}
\pi_i \in \{0,1\}^{P},\qquad
\pi_i[(j,\ell)] = \mathbb{I}\{(j,\ell)\rightarrow i\},
\end{equation}
Here, $P=D\tau_{\max}$ indexes all the sensors and the lag pairs.
This mask identifies allowable parent features and is applied as a hard gate to the lagged input. 
As a result, only the graph-supported parent features can contribute to the causal computation path, which includes the Transformer encoder, latent variable module, and the causal predictive head.

\subsection{Target-Specific Forecasting Blocks}
\label{sec:blocks}
\begin{figure*}[t!]
    \centering
    \includegraphics[width=\textwidth]{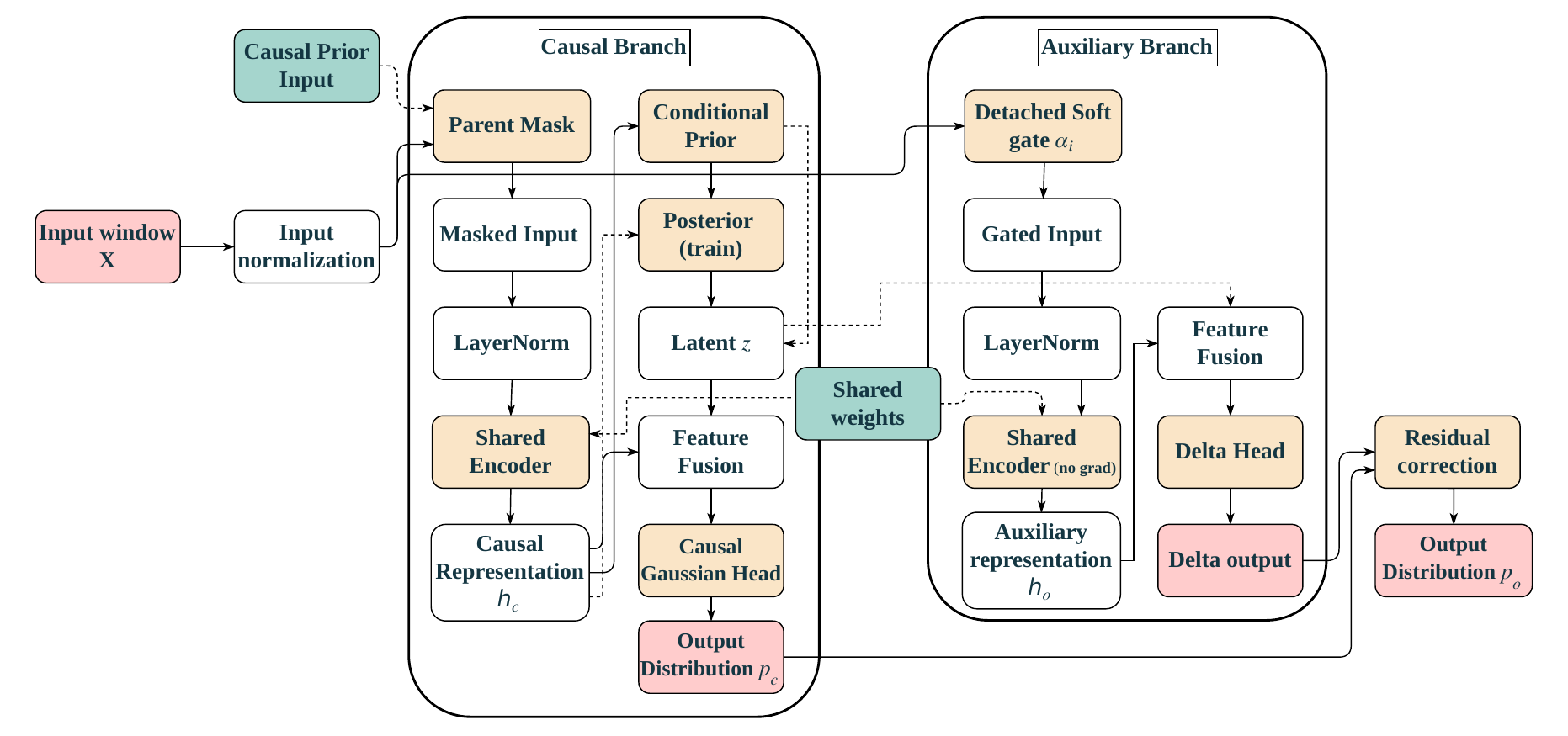}
    \caption{Detailed architecture of a target-specific forecasting block. The causal branch applies the parent mask and shared Transformer encoder to produce a causal representation and a latent probabilistic prediction. The auxiliary branch uses a detached soft gate and no-gradient encoder evaluation to learn residual corrections without updating the causal modules.}
    \label{fig:t_block}
\end{figure*}

Each target variable $i \in \{1,\dots,D\}$ is assigned a dedicated forecasting block with its own parameters, which includes a target-specific parent mask $\pi_i$ derived from the causal graph and a set of learnable auxiliary gate logits. During training and evaluation, mini-batches are constructed so that all samples in a batch have the same target variable. 
This deterministic routing enables each batch to be processed by its corresponding target-specific block and simplifies block-wise computation. Figure~\ref{fig:t_block} illustrates the internal structure of a single target-specific forecasting block. The parent mask \(\pi_i\) defines the causal input pathway, which produces the causal density \(p_c^{(i)}\). The detached soft gate \(\alpha_i\) defines the residual auxiliary pathway, which produces the auxiliary density \(p_o^{(i)}\). These two densities are later converted into the causal and auxiliary NLL scores \(s_{t,i}^{(c)}\) and \(s_{t,i}^{(o)}\), which are combined during inference.

\subsection{Causal Transformer Encoder and Latent Gaussian Head}
\label{sec:causal}
For each target $i$, the lagged input matrix $X_t \in \mathbb{R}^{W \times P}$ is first restricted to graph-supported parent features using the binary parent mask $\pi_i \in \{0,1\}^{P}$:
\begin{equation}
X^{(i)}_{t,c} = X_t \odot \pi_i.
\end{equation}

The masked input is then processed by two operations before being passed to the Transformer encoder. First, \(\mathrm{LN}(\cdot)\) applies layer normalization independently to each time token over the feature dimension. Second, a linear projection maps each normalized \(P\)-dimensional lag-feature token into the Transformer model dimension \(d\):
\begin{equation}
\tilde{X}^{(i)}_{t,c} = \mathrm{LN}\!\left(X^{(i)}_{t,c}\right),
\qquad
E_t = \tilde{X}^{(i)}_{t,c} W_p + b_p,
\end{equation}
where \(W_p \in \mathbb{R}^{P \times d}\), \(b_p \in \mathbb{R}^{d}\), and \(E_t \in \mathbb{R}^{W \times d}\). Thus, each row of \(E_t\) is a \(d\)-dimensional embedding of the corresponding normalized lagged input token.

The projected tokens are processed by a Transformer encoder:
\begin{equation}
H_t = \mathrm{TransEnc}(E_t) \in \mathbb{R}^{W \times d},
\end{equation}
and the final token representation is used as the causal summary
\begin{equation}
h^{(i)}_{t,c} = H_t[W-1] \in \mathbb{R}^{d}.
\end{equation}

To capture residual uncertainty after conditioning on the causal representation, a latent variable $z \in \mathbb{R}^{d_z}$ is introduced. The model defines a conditional Gaussian prior
\begin{equation}
p_{\psi}(z\mid h_{t,c})
= \mathcal{N}\!\left(\mu_p(h_{t,c}),\ \mathrm{diag}\!\big(\exp(\log v_p(h_{t,c}))\big)\right),
\end{equation}
and, during training, an amortized posterior conditioned on both the causal representation and the ground-truth target \(y_{t,i}=x_t^{(i)}\):
\begin{equation}
q_{\phi}(z\mid h_{t,c},y)
= \mathcal{N}\!\left(\mu_q(h_{t,c},y),\ \mathrm{diag}\!\big(\exp(\log v_q(h_{t,c},y))\big)\right).
\end{equation}
Samples are drawn using the reparameterization trick,
\begin{equation}
z=\mu_q + \exp\!\left(\tfrac{1}{2}\log v_q\right)\odot\epsilon,
\qquad \epsilon\sim\mathcal{N}(0,I),
\end{equation}
and the latent module is regularized by
\begin{equation}
\mathcal{L}_{\mathrm{KL}}
= \mathrm{KL}\!\left(q_{\phi}(z\mid h_{t,c},y)\ \|\ p_{\psi}(z\mid h_{t,c})\right).
\end{equation}

The causal representation and latent sample are concatenated,
\begin{equation}
u_{t,c} = [h_{t,c}; z],
\end{equation}
and mapped to the parameters of a Gaussian predictive distribution:
\begin{equation}
\mu_c = g_{\mu}(u_{t,c}),
\qquad
\log v_c = g_{v}(u_{t,c}).
\end{equation}
This yields
\begin{equation} 
p_c^{(i)}(y_{t,i}\mid X_t,\pi_i,z) = \mathcal{N}\!\left(y_{t,i};\mu_c^{(i)},\exp(\log v_c^{(i)})\right), 
\end{equation}
with per-sample causal negative log-likelihood
\begin{equation}
\mathcal{L}_c
= \frac{1}{2}\left(\log(2\pi)+\log v_c + \frac{(y-\mu_c)^2}{\exp(\log v_c)}\right).
\end{equation}
Only this causal branch updates the Transformer trunk, latent module, and causal Gaussian head. During inference, the latent variable is sampled from the conditional prior $p_{\psi}(z\mid h_{t,c})$.


\subsection{Gradient-Isolated Residual Auxiliary Path}
\label{sec:shadow}
The auxiliary path is designed to exploit additional correlations, including non-parent inputs, without modifying the causal trunk, latent module, or causal prediction head through gradient flow. A learnable soft gate is defined as
\begin{equation}
\alpha_i = \sigma(\ell_i) \in (0,1)^P,
\end{equation}
where $\ell_i \in \mathbb{R}^{P}$ denotes target-specific gate logits. Let $\sg(\cdot)$ denote the stop-gradient operator. The auxiliary branch receives a detached version of the gate:
\begin{equation}
X^{(i)}_{t,o} = X_t \odot \sg(\alpha_i).
\end{equation}
The same Transformer trunk is reused to encode the auxiliary input, but it is evaluated without gradient tracking:
\begin{equation}
h^{(i)}_{t,o} = \sg\!\left(f_{\theta}\!\left(\mathrm{LN}\!\left(X^{(i)}_{t,o}\right)\right)\right).
\end{equation}
The auxiliary branch also receives a detached latent sample. Its fused representation is
\begin{equation}
u_{t,o} = \big[h^{(i)}_{t,o}; \sg(z)\big].
\end{equation}
Residual corrections to the causal Gaussian parameters are then predicted by
\begin{equation}
\begin{aligned}
\Delta\mu &= g_{\Delta\mu}(u_{t,o}),\\
\Delta\log v &= g_{\Delta v}(u_{t,o}).
\end{aligned}
\end{equation}
The corrected predictive parameters are defined as
\begin{equation}
\mu_o = \mu_c + \Delta\mu,
\qquad
\log v_o = \log v_c + \Delta\log v.
\end{equation}
The corresponding residual auxiliary predictive density is defined as \begin{equation} p_o^{(i)}(y_{t,i}\mid X_t,\alpha_i,z) = \mathcal{N}\!\left(y_{t,i};\mu_o^{(i)},\exp(\log v_o^{(i)})\right). \end{equation} Here, \(p_o^{(i)}\) does not represent an independent forecasting model. Instead, it denotes the Gaussian density obtained after applying residual corrections to the causal predictive parameters.

During training, the auxiliary likelihood is evaluated using a detached causal base so that auxiliary gradients do not propagate into the causal branch:
\begin{equation}
\begin{aligned}
\tilde{\mu}_o &= \sg(\mu_c) + \Delta\mu,\\
\widetilde{\log v}_o &= \sg(\log v_c) + \Delta\log v.
\end{aligned}
\end{equation}
The auxiliary negative log-likelihood is
\begin{equation}
\mathcal{L}_o
= -\log \mathcal{N}\!\left(y;\tilde{\mu}_o,\exp(\widetilde{\log v}_o)\right).
\end{equation}
As a result, $\mathcal{L}_o$ updates only the auxiliary residual-prediction layers, while the gate parameters are learned exclusively through the regularization terms described below.

To discourage large auxiliary overrides, the residual magnitudes are penalized:
\begin{equation}
\mathcal{L}_{\mathrm{res}}
= \mathbb{E}\!\left[\Delta\mu^{2}\right] + \mathbb{E}\!\left[(\Delta\log v)^{2}\right].
\end{equation}
\subsection{Gate Regularization and Causal Biasing of $\alpha$}
\label{sec:gate_reg}
Because $\alpha$ may include non-parent inputs, regularization is applied to encourage a sparse structure primarily composed of parent inputs.
Let $\alpha=\sigma(\ell)$ represent the soft gate induced by logits $\ell \in \mathbb{R}^{P}$.

\subsubsection{Parent-alignment prior}
The logits are biased toward the hard mask $\pi_i$ through the use of weighted binary cross-entropy:
\begin{equation}
\mathcal{R}_{\mathrm{prior}}
= \lambda_{\mathrm{prior}}\ \mathrm{BCEWithLogits}(\ell,\pi_i;w),
\end{equation}
Here, $w$ assigns greater weight to parent indices, thereby promoting stable inclusion of graph-supported parents.

\subsubsection{Asymmetric sparsity and separation margin}
Gate mass on non-parents is penalized while high gate values are encouraged for parents:
\begin{equation}
\mathcal{R}_{\mathrm{other}}=\lambda_{\mathrm{other}}\ \mathbb{E}[\alpha\mid \pi_i=0],\quad
\mathcal{R}_{\mathrm{push}}=\lambda_{\mathrm{push}}\ \mathbb{E}[1-\alpha\mid \pi_i=1].
\end{equation}
A soft margin is introduced to promote separation between the average gate values of parents and non-parents:
\begin{equation}
\mathcal{R}_{\mathrm{margin}}
= \lambda_m \max\!\left(0,\bar{\alpha}_{\mathrm{oth}}-\bar{\alpha}_{\mathrm{par}}+m\right).
\end{equation}

\subsubsection{Group penalty across lags per sensor}
Gate entries are grouped by sensor across all lags, and a group penalty is imposed:
\begin{equation}
\mathcal{R}_{\mathrm{grp}}
= \lambda_{\mathrm{grp}}\,\frac{1}{D}\sum_{j=1}^{D}\|\alpha_{G_j}\|_2,
\end{equation}
This approach encourages sensor-level sparsity, or group-sparsity, across lagged features \cite{yuan2006grouplasso}.

\subsection{Training Objective and Optimization}
\label{sec:training}
For each target $i$, the training procedure minimizes a weighted sum of the causal and auxiliary NLLs, in addition to regularization terms:
\begin{equation}
\mathcal{L}
= (1-\gamma_t)\mathcal{L}_c + \gamma_t \mathcal{L}_o
+ \beta_t \mathcal{L}_{\mathrm{KL}}
+ \lambda_{\mathrm{res}}\mathcal{L}_{\mathrm{res}}
+ \mathcal{R}(\alpha),
\end{equation}
where $\mathcal{R}(\alpha)=\mathcal{R}_{\mathrm{prior}}+\mathcal{R}_{\mathrm{other}}+\mathcal{R}_{\mathrm{push}}+\mathcal{R}_{\mathrm{margin}}+\mathcal{R}_{\mathrm{grp}}$.
In accordance with the implementation, the shadow weight $\gamma_t$ and KL weight $\beta_t$ are increased linearly during the initial $E_{\mathrm{warm}}$ epochs:
\begin{equation}
\gamma_t=\min\!\left(\gamma,\ \frac{e+1}{E_{\mathrm{warm}}}\gamma\right),\qquad
\beta_t=\beta\min\!\left(1,\ \frac{e+1}{E_{\mathrm{warm}}}\right),
\end{equation}
where $e$ is the epoch index (starting at $0$).
Optimization is performed using the Adam algorithm with gradient-norm clipping.
To ensure numerical stability, log-variance parameters are constrained within bounded intervals, and non-finite losses are excluded during training.

\subsection{Inference: NLL, Blended Scoring, and Aggregation}
\label{sec:scoring}
For each time point $t$ and target $i$, the causal and auxiliary NLLs are computed
\begin{equation} 
\begin{aligned} 
s_{t,i}^{(c)} &= -\log p_c^{(i)}(y_{t,i}\mid X_t,\pi_i,z),\\
s_{t,i}^{(o)} &= -\log p_o^{(i)}(y_{t,i}\mid X_t,\alpha_i,z). 
\end{aligned} 
\end{equation}
When $S$ samples of the latent variable $z$ are used, NLLs are computed for each sample and averaged across $S$ samples.
A blended per-target score is subsequently defined as
\begin{equation}
s_{t,i}(\gamma) = (1-\gamma)s_{t,i}^{(c)} + \gamma s_{t,i}^{(o)}.
\end{equation}
Aggregating across targets produces a scalar anomaly score
\begin{equation}
S_t = \Agg\Big(\{s_{t,i}(\gamma)\}_{i=1}^{D}\Big),
\end{equation}
where $\Agg(\cdot)$ is implemented as the mean, maximum, or top-$k$ mean.

\subsection{Safe Activation (Selection of $\gamma$)}
\label{sec:safety}
To mitigate unsafe reliance on non-parent correlations, an effective blend weight $\gamma_{\mathrm{used}}$ is selected at deployment time.
The safety statistics are computed on an unlabeled calibration prefix of the deployment stream, after which $\gamma_{\mathrm{used}}$ is fixed for subsequent scoring.
(i) Blended scores $S_t^{\mathrm{mix}}$ are computed under a base $\gamma$, and (ii) stressed scores $S_t^{\mathrm{permNP}}$ are obtained after permuting non-parent input features across samples within each target-specific scoring batch. This procedure preserves feature marginals while disrupting sample-wise alignment.
Sensitivity is quantified by
\begin{equation}
R = \frac{\mathbb{E}_t\!\left[\left|S_t^{\mathrm{mix}}-S_t^{\mathrm{permNP}}\right|\right]}{\mathbb{E}_t\!\left[\left|S_t^{\mathrm{mix}}\right|\right]+\epsilon}.
\end{equation}
The permutation-sensitivity diagnostic is not intended to provide a formal proof of causal safety. Instead, it is used as an operational model-level test of whether the blended score is overly dependent on non-parent inputs. If the auxiliary residual path relies strongly on non-parent correlations, perturbing non-parent lagged features while keeping graph-supported parent features unchanged can substantially change the blended anomaly score. Conversely, if the score is mainly determined by the graph-supported parent pathway, it should remain relatively stable under this perturbation. Therefore, a high sensitivity value indicates possible reliance on unstable or potentially non-causal correlations, and the effective auxiliary blending weight \(\gamma_{\mathrm{used}}\) is reduced. This diagnostic should be interpreted as a practical robustness check relative to the learned graph prior, rather than as a formal guarantee of physical causal validity.
Gate separation is also measured:
\begin{equation}
M = \mathbb{E}_i\!\left[\bar{\alpha}_{\mathrm{par}}^{(i)}-\bar{\alpha}_{\mathrm{oth}}^{(i)}\right].
\end{equation}
If $R>\tau_{\mathrm{rel}}$ or $M<\tau_{\alpha}$, auxiliary influence is reduced by setting $\gamma_{\mathrm{used}}=0$ hard fallback or by soft-scaling
\begin{equation}
\gamma_{\mathrm{used}} = \gamma \cdot \mathrm{clip}\!\left(\frac{M-\tau_{\alpha}}{1-\tau_{\alpha}},\,0,\,1\right),
\end{equation}
otherwise $\gamma_{\mathrm{used}}=\gamma$.
The selected $\gamma_{\mathrm{used}}$ remains fixed during inference scoring.

\subsection{Thresholding with Streaming Peaks-Over-Threshold}
\label{sec:spot}
Given the aggregated anomaly score sequence $\{S_t\}_{t=1}^{T}$, scores are converted into binary anomaly decisions using SPOT (Streaming Peaks-Over-Threshold), an EVT-based streaming thresholding procedure \cite{siffer2017spot}.
SPOT follows the Peaks-Over-Threshold (POT) principle. Any value exceeding a high threshold is modelled using extreme value theory, via a Generalised Pareto tail approximation, to produce an adaptive decision threshold in non-stationary streams \cite{coles2001extremes}.

\paragraph{Dynamic SPOT thresholds.}
SPOT is configured by a risk tail probability parameter $q$ and an initialization quantile level. Let $(s_1,\ldots,s_N)$ denote the anomaly-score sequence
after the initial window-and-lag exclusion. We define
$\mathcal I=(s_1,\ldots,s_b)$ as the initialization
prefix used to initialize SPOT, and
$\mathcal R=(s_{b+1},\ldots,s_N)$ as the subsequent
sequence used for thresholded anomaly detection and the initialization prefix is excluded from the
reported thresholded detection metrics.q
SPOT is fitted on $\mathcal{I}$ and subsequently run in streaming mode on $\mathcal{R}$ to produce a \emph{time-varying} threshold sequence $\{\theta_t\}$:
\begin{equation}
\theta_t = \mathrm{SPOT}\big(\mathcal{I},\mathcal{R};q,\text{level}\big)\quad \text{(dynamic update)}.
\end{equation}
A scalar calibration factor $\lambda_{\mathrm{thr}}$ is applied to adjust the level of conservativeness:
\begin{equation}
\tilde{\theta}_t = \lambda_{\mathrm{thr}}\,\theta_t.
\end{equation}

\paragraph{Point decisions and segment-level adjustment.}
A raw pointwise prediction is obtained by applying the threshold:
\begin{equation}
\hat{y}_t^{\mathrm{raw}} = \mathbb{I}\{S_t^{+} > \tilde{\theta}_t\},\qquad t \in \mathcal{R}.
\end{equation}

\subsection{Root-Cause Variable Attribution}
\label{sec:rca}
In addition to generating a scalar anomaly score $S_t$, the method provides a ranked list of sensors most responsible for an anomalous event at time $t$.
Attribution scores are derived from the same probabilistic forecasting model used for detection.
These attributions are model-based: they quantify how the detector's likelihood and anomaly score change under input replacement, and do not constitute guaranteed physical causal interventions.

\subsubsection{Per-dimension blended NLL attribution}
For each time $t$ and variable $i$, a blended per-variable negative log-likelihood is computed:
\begin{equation}
A_{t,i} = (1-\gamma_{\mathrm{used}})\,s^{(c)}_{t,i} + \gamma_{\mathrm{used}}\,s^{(o)}_{t,i},
\end{equation}
where $s^{(c)}_{t,i}$ and $s^{(o)}_{t,i}$ are the causal and auxiliary NLLs, respectively.
To ensure comparability across variables, each dimension is standardized using a nominal baseline prefix to estimate $\mu_i$ and $\sigma_i$:
\begin{equation}
Z_{t,i} = \frac{A_{t,i}-\mu_i}{\sigma_i+\varepsilon}.
\end{equation}
Variables are ranked by $Z_{t,i}$ at anomalous times, where higher values indicate a stronger contribution.

\subsubsection{Counterfactual-Style Input Clamping Attribution}
Variable contributions are also quantified via model-based counterfactual-style input clamping.
For a candidate sensor $s$, only the historical inputs corresponding to $s$ across all lags within the window are replaced with a nominal reference value, the training median of sensor $s$, and the anomaly score is recomputed:
\begin{equation}
\Delta_s(t) = S_t^{\mathrm{clamp}(s)} - S_t^{\mathrm{orig}}.
\end{equation}
A large negative $\Delta_s(t)$ indicates that clamping sensor $s$ substantially reduces the anomaly score, thereby suggesting $s$ as a root cause.
For computational efficiency, only target blocks that are potentially influenced by $s$ are recomputed using the causal parent mask and non-negligible auxiliary gates.

\section{Experimental Results}
\label{sec:experiments}
This section describes the evaluation of the proposed CGT for anomaly detection and root-cause variable attribution on multivariate time series. It presents comparison results on two benchmark datasets and reports ablation studies that evaluate the effectiveness of each module. 

\subsection{Datasets}
\label{subsec:datasets}
Evaluation is performed using three multivariate time series benchmarks that provide timestamp-level anomaly annotations. Table \ref{tab:datasets} shows the datasets used in this study. Both ASD and SMD contain 12 entities with dimension-level labels. The SMD has a higher feature dimensionality and better sampling resolution than ASD. Additionally, SWaT dataset used because it captures realistic process behavior under both normal and attack conditions and it is widely used to evaluate anomaly detection methods for industrial and cyber-physical systems.

\subsubsection{Server Machine Dataset (SMD)}
\label{subsubsec:smd}
The first public benchmark that we used is SMD dataset ~\cite{su2019robust}, which contains $12$ entities, and each with $D=38$ metrics sampled uniformly every one minute. SMD provides timestamp-level anomaly labels and variable annotations used for attribution evaluation.

\subsubsection{Application Server Dataset (ASD)}
\label{subsubsec:asd}
The second dataset used in this study is the ASD dataset \cite{li2021interfusion}, which contains $12$ servers, each represented by $D=19$ metrics covering CPU, memory, network, and virtual machine signals. Samples are recorded based on five-minute intervals. According to the dataset explanation, the initial 30 days are allocated for training, with the final 30\% of this period reserved for validation. The subsequent 15 days constitute the test set. Test anomalies and their most anomalous variables are annotated by domain experts for evaluation.

\subsubsection{Secure Water Treatment (SWaT)}
\label{subsubsec:swat}
The SWaT dataset was introduced by ~\cite{mathur2016swat} as a water treatment testbed for research and training in industrial control system security. It closely resembles the operation of a real water treatment plant and contains multivariate time-series data collected over 7 days of normal operation and 4 days of abnormal operation. The dataset includes sensor readings, such as water level, flow rate, pressure, and conductivity, as well as actuator states for components such as pumps and valves. 

\begin{table}[t] \centering \caption{Dataset characteristics used in the evaluation.} \label{tab:datasets} \footnotesize \setlength{\tabcolsep}{3pt} \renewcommand{\arraystretch}{1.05} \begin{tabular}{lcccccc} \hline Dataset & Ent. & $D$ & Samp. & Train & Test & Dim. labels \\ \hline ASD & 12 & 19 & 5m & 30d & 15d & Yes \\ SMD & 12 & 38 & 1m & first half & second half & Yes \\ SWaT & 1 & 51 & 1s & 7d normal & 4d attacks & No \\ \hline \end{tabular} \end{table}

\subsection{Baselines}
\label{subsec:baselines}
This study compared the CGT against representative state-of-the-art forecasting- and reconstruction-based anomaly detectors:
OmniAnomaly~\cite{su2019robust}, MSCRED~\cite{zhang2019mscred}, MAD-GAN~\cite{li2019madgan}, USAD~\cite{audibert2020usad}, MTAD-GAT~\cite{zhao2020mtadgat}, CAE-M~\cite{zhang2021unsupervised}, GDN~\cite{deng2021graph}, and TranAD~\cite{tuli2022tranad}.
Each baseline is implemented based on the publicly released code. 

\subsection{Preprocessing and Feature Construction}
\label{subsec:data_preproc}
All features are normalised using Min-Max scaling and then fitted to the training split; this transformation is subsequently applied to the test data without modification. For each timestamp $t$ and target variable $i$, we construct the windowed lag input $X_t\in\mathbb{R}^{W\times P}$ using window length $W$ and maximum lag $\tau_{\max}$, where $P=D\tau_{\max}$.
Unless otherwise specified, $W=30$ and $\tau_{\max}=7$ are used.
Scores and predictions are defined only for $t \ge W+\tau_{\max}$.

\subsection{Causal Graph Estimation (PCMCI Prior)}
\label{subsec:pcmci}
The parent mask $\pi_i$ for each target $i$ is from estimation of a time-lagged causal graph on the \emph{training split only} using a PCMCI-style causal discovery procedure.
Directed edges $(j,\ell)\rightarrow i$ for lags $\ell\in\{1,\dots,\tau_{\max}\}$ are mapped to binary mask entries $\pi_i[(j,\ell)]\in\{0,1\}$, and they are used to hard-mask non-parent lag-features in the causal path. The resulting PCMCI graph is not used as ground-truth causality; it is used as a sparse structural prior for constructing the parent masks \(\pi_i\) used by the causal forecasting branch.

\subsection{Evaluation Protocol}
\label{subsec:evaluation}
CGT produces a continuous anomaly score $S_t$; instead of binary labels, it uses an aggregated negative log-likelihood. Detection performance is evaluated by thresholding $S_t$ to obtain binary predictions.

\textbf{Thresholding}
We use SPOT (Streaming Peaks-Over-Threshold)~\cite{siffer2017spot} to obtain dynamic thresholds on the test stream.

\textbf{Detection metrics.}
Precision, Recall, and F1-score are evaluation metrics. 

AUROC is also reported, computed from the anomaly scores to assess the quality of the ranking.

\subsection{Implementation Details}
\label{subsec:impl_details}
All experiments are implemented in PyTorch with fixed random seeds to ensure reproducibility. CGT is implemented in PyTorch. Table~\ref{tab:hparams} report the
dataset specific architecture, optimization and calibration settings and 
the ASD and SMD configuration uses an input window of 30 observations,
whereas SWaT uses 12 and both use a maximum lag of seven observations.
The SWaT model use smaller feed-forward and latent dimensions. For SWaT, hyperparameters were selected using only normal-operation
data, since no labeled validation split is available. A held-out
normal segment was used for model selection and calibration, while
attack labels were reserved exclusively for final evaluation.

The multiplier $\lambda_{\mathrm{thr}}$ scales the anomaly-detection
threshold. A value of 1.0 leaves the threshold unchanged.
The thresholds $\tau_{\mathrm{rel}}$ and $\tau_{\alpha}$ specify the
non-parent sensitivity and gate-separation criteria used by the
safety mechanism. Auxiliary influence is reduced when
$R>\tau_{\mathrm{rel}}$ or $M<\tau_{\alpha}$.
The base weight $\gamma_{\mathrm{base}}$ is the auxiliary blending
coefficient before this attenuation and it correspond to $\gamma$ in
the score-blending equations. These configured parameters are
distinct from the observed diagnostics $R$, $M$, and the effective
weight $\gamma_{\mathrm{used}}$ reported in the safety-diagnostics table.

\begin{table}[t]
\centering
\caption{Dataset-specific CGT architecture, optimization and calibration settings.}
\label{tab:hparams}
\small\renewcommand{\arraystretch}{1.12}
\begin{tabular}{lcc}
\toprule
Hyperparameter & ASD-SMD & SWaT\\
\midrule
Window $W$ &30&12\\
Maximum lag $\tau_{\max}$ &7&7\\
Model dimension $d_{\mathrm{model}}$ &64&64\\
Attention heads &2&2\\
Encoder layers &2&2\\
Feed-forward dimension $d_{\mathrm{ff}}$ &128&48\\
Latent dimension $d_z$ &8&6\\
Batch size &16&64\\
Learning rate & $0.00039$ & $0.0001$\\
Gradient clipping &0.82259&0.88234\\
\midrule
Threshold multiplier $\lambda_{\mathrm{thr}}$ &1.0&1.0\\
Sensitivity threshold $\tau_{\mathrm{rel}}$ &0.0862778&0.713365\\
Gate-margin threshold $\tau_{\alpha}$ &0.0365172&0.043565\\
Base auxiliary weight $\gamma_{\mathrm{base}}$ &0.0206281&0.0197331\\
\bottomrule
\end{tabular}
\end{table}

\subsection{Results and Discussion}
\label{subsec:results}
Table~\ref{tab:main_detection} reports the precision, recall, F1, and AUC scores for baseline models and CGT on the SMD, ASD, and SWaT datasets. On SMD, CGT achieves the highest F1-score of 95.32\%, demonstrating a strong balance between precision and recall. On ASD, CGT also outperforms the baselines with the best F1-score of 96.19\% and the highest AUROC of 99.11\%, indicating improved detection reliability. On SWaT, CGT further achieves the best F1-score of 83.91\% and the highest AUROC of 86.28\%, showing that the proposed method generalizes effectively to industrial data. Overall, the results indicate that CGT improves anomaly detection performance across diverse multivariate time-series benchmarks by maintaining competitive precision while enhancing recall and F1-score. Although the improvement over some strong baselines is moderate, CGT provides additional diagnostic functionality that these baselines do not explicitly offer. In particular, CGT combines likelihood-based anomaly scoring with a graph-guided forecasting pathway and model-based attribution, allowing detected anomalies to be linked to target-specific lagged dependencies. Therefore, the contribution of CGT should be interpreted not only in terms of absolute F1-score gains, but also in terms of the improved balance between detection performance, uncertainty-aware scoring, and interpretability.

\begin{table*}[t]
\centering
\caption{Anomaly detection results in (\%) on SMD, ASD, and SWaT.}
\label{tab:main_detection}
\footnotesize
\setlength{\tabcolsep}{3pt}
\renewcommand{\arraystretch}{1.05}
\begin{tabular}{l|cccc|cccc|cccc}
\hline
\multirow{2}{*}{Method}
& \multicolumn{4}{c|}{SMD}
& \multicolumn{4}{c|}{ASD}
& \multicolumn{4}{c}{SWaT} \\
& P & R & F1 & AUROC & P & R & F1 & AUROC & P & R & F1 & AUROC \\
\hline
OmniAnomaly~\cite{su2019robust} & 88.74 & \textbf{99.79} & 93.94 & 99.41 & 78.36 & 99.18 & 87.55 & 87.59 & 97.69 & 69.72 & 81.24 & 84.59 \\
MSCRED~\cite{zhang2019mscred} & 72.68 & 99.68 & 84.07 & 99.17 & 89.01 & 98.54 & 93.53 & 98.01 & \textbf{99.86} & 67.84 & 80.86 & 84.41 \\
MAD-GAN~\cite{li2019madgan} & \textbf{99.87} & 84.32 & 91.44 & 99.28 & 85.08 & 99.21 & 91.60 & 98.56 & 96.07 & 69.39 & 80.52 & 84.55 \\
USAD~\cite{audibert2020usad} & 90.51 & 99.69 & 94.88 & 99.29 & 79.38 & 99.06 & 88.13 & 97.89 & 99.64 & 68.91 & 81.37 & 84.72 \\
MTAD-GAT~\cite{zhao2020mtadgat} & 82.01 & 92.07 & 86.75 & 99.18 & 79.09 & 98.17 & 87.60 & 98.92 & 97.31 & 69.69 & 81.18 & 84.58 \\
CAE-M~\cite{zhang2021unsupervised} & 90.73 & 96.64 & 93.59 & 97.78 & 77.42 & \textbf{99.92} & 87.24 & 98.98 & 96.84 & 69.43 & 80.93 & 84.70 \\
GDN~\cite{deng2021graph} & 71.62 & 99.67 & 83.35 & 99.19 & 92.99 & 98.85 & 95.83 & 98.09 & 97.12 & 69.70 & 81.14 & 84.51 \\
TranAD~\cite{tuli2022tranad} & 94.75 & 94.65 & 94.70 & \textbf{99.54} & 88.58 & 90.27 & 89.42 & 98.46 & 97.74 & 70.11 & 81.64 & 85.02 \\
\hline
CGT & 91.26 & 99.75 & \textbf{95.32} & 99.38 & \textbf{93.48} & 99.07 & \textbf{96.19} & \textbf{99.11} & 98.27 & \textbf{73.21} & \textbf{83.91} & \textbf{86.28} \\
\hline
\end{tabular}
\end{table*}

\subsubsection{Ablation}

To clarify the role of each ablation, we explicitly map each experimental variant to the corresponding CGT component. A0 tests the graph-constrained causal forecasting branch alone. A1 tests the contribution of the residual auxiliary path by adding it with a fixed blending weight. A2 tests the proposed safety-gated blending mechanism, which selectively controls the auxiliary contribution during inference. The safety-gate diagnostics further analyze the reliability mechanism used in A2, while the root-cause attribution experiments evaluate the diagnostic module based on per-dimension likelihood scores and model-based input clamping. 

\paragraph{Scoring Variants}
We isolate the inference-time scoring effects by evaluating A0, A1, and A2 on the SMD dataset using the same trained checkpoint and identical threshold settings. Table~\ref{tab:ablation_scoring} shows the comparison of three ablation settings of the scoring method, the way the model outputs are combined to produce the score during testing. The scoring ablation follows the formal definitions in Section~\ref{sec:scoring}. A0 is the causal-only setting and uses only the causal NLL: \[ s_{t,i}^{\mathrm{A0}} = s_{t,i}^{(c)}. \] A1 uses a fixed residual auxiliary blend with the base value \(\gamma_{\mathrm{base}}\): \[ s_{t,i}^{\mathrm{A1}} = (1-\gamma_{\mathrm{base}})s_{t,i}^{(c)} + \gamma_{\mathrm{base}}s_{t,i}^{(o)}. \] A2 is the proposed safety-gated setting. It uses the same blended scoring rule, but replaces \(\gamma_{\mathrm{base}}\) with the effective value \(\gamma_{\mathrm{used}}\) selected by the safety diagnostics: \[ s_{t,i}^{\mathrm{A2}} = (1-\gamma_{\mathrm{used}})s_{t,i}^{(c)} + \gamma_{\mathrm{used}}s_{t,i}^{(o)}. \] Thus, A0 evaluates the graph-constrained causal branch alone, A1 evaluates unconditional use of the residual auxiliary path, and A2 evaluates the proposed selective use of residual auxiliary information.
It shows that adding the residual branch in A1 slightly improves precision, recall, F1, and AUROC compared to the causal-only baseline. The A2, the third setting, which is the proposed safety gate mechanism, achieves the better results on all metrics, which indicate that the residual information is effective when it is applied selectively using the safety gate rather than always being blended in. 

\begin{table}[t]
\centering
\caption{Scoring ablation results in (\%) on the SMD dataset. A0 uses
causal-only negative log-likelihood scoring, A1 adds the residual
auxiliary path with a fixed blending weight, and A2 applies the
proposed safety-gated blending mechanism. Higher values indicate better performance.}
\label{tab:ablation_scoring}
\footnotesize
\setlength{\tabcolsep}{4pt}
\renewcommand{\arraystretch}{1.05}
\begin{tabular}{lcccc}
\hline
Variant & P & R & F1 & AUROC \\
\hline
A0 (Causal-only)        & 90.82 & 99.11 & 94.88 & 98.94 \\
A1 (Fixed blend)        & 91.05 & 99.38 & 95.10 & 99.12 \\
A2 (Safety-gated blend) & \textbf{91.26} & \textbf{99.75} & \textbf{95.32} & \textbf{99.38} \\
\hline
\end{tabular}
\end{table}

\paragraph{Safety-gate behavior}
We report safety diagnostics to show when and how strongly the gate suppresses auxiliary scoring. Table~\ref{tab:safety_diagnostics} shows that the safety gate does not improve performance simply by chance; instead, it reduces the residual branch when the causal alignment is weaker during the testing. The value $R$ measures how sensitive the model is to non-parent features, while $M$ measures how much more strongly the gate favors parent features over non-parent features. In both datasets, $M$ is positive, indicating that the gate places greater importance on parent features. SMD has a higher $R$ and a lower $M$ than ASD, suggesting that non-parent effects are somewhat stronger and the gate is less clearly separated on SMD. This shows that the gate acts as intended by being more conservative when the residual branch appears less reliable, leading to a lower $\gamma_{\mathrm{used}}$ and a higher fallback rate. This supports the results in Table~\ref{tab:ablation_scoring}, showing that the residual branch can be helpful but should be used carefully when it is less consistent with the causal structure.

\begin{table}[t]
\centering
\caption{Inference-time safety diagnostics for the A2 scoring variant on the ASD and SMD datasets. \(R\) denotes the relative sensitivity of the blended anomaly score to permutation of non-parent inputs, and \(M\) denotes the average parent--non-parent gate-separation margin.}
\label{tab:safety_diagnostics}
\footnotesize
\setlength{\tabcolsep}{4pt}
\renewcommand{\arraystretch}{1.05}
\begin{tabular}{lccccc}
\hline
Dataset & $R$ & $M$ & $\gamma_{\mathrm{base}}$ & $\gamma_{\mathrm{used}}$ & Fallback (\%) \\
\hline
ASD & 0.0437 & 0.0814 & 0.0206 & 0.0191 & 14.3 \\
SMD & 0.0612 & 0.0678 & 0.0206 & 0.0174 & 28.6 \\
\hline
\end{tabular}
\end{table}
In Table~\ref{tab:safety_diagnostics}, Fallback (\%) denotes how often the safety mechanism determines that the auxiliary residual path should not be fully trusted. A fallback event occurs when the non-parent sensitivity \(R\) exceeds the safety threshold or when the gate margin \(M\) falls below the required margin. In such cases, the effective auxiliary weight \(\gamma_{\mathrm{used}}\) is reduced relative to \(\gamma_{\mathrm{base}}\), or set to zero in the hard-fallback case, causing the final anomaly score to rely more strongly on the causal branch.
\paragraph{Root-cause variable attribution}
This section reports two attribution variants: the counterfactual-style clamp score drop and the per-dimension standardised blended NLL ($z$-score). Higher values indicate better attribution performance. For root-cause attribution, we report Hit Rate and normalized discounted cumulative gain (NDCG). Let \(G_t\) denote the set of ground-truth anomalous variables at anomalous timestamp \(t\), and let \(\hat{R}_t\) be the ranked list of variables produced by an attribution method. For a percentage \(P \in \{100,150\}\), the cutoff is defined as
\begin{equation}
k_t(P)=\max\left(1,\mathrm{round}\left(\frac{P|G_t|}{100}\right)\right).
\end{equation}
H@\(P\)\% denotes Hit Rate@\(P\)\%, for example, the fraction of ground-truth anomalous variables recovered within the top \(k_t(P)\) ranked variables:
\begin{equation}
\mathrm{H@}P\% =
\frac{1}{|\mathcal{T}_a|}
\sum_{t\in\mathcal{T}_a}
\frac{\left|G_t\cap \hat{R}_t[:k_t(P)]\right|}{|G_t|},
\end{equation}
where \(\mathcal{T}_a\) is the set of evaluated anomalous timestamps. N@\(P\)\% denotes NDCG@\(P\)\% computed using the same cutoff \(k_t(P)\) and binary relevance labels for the ground-truth anomalous variables. For example, if two variables are labeled anomalous at timestamp \(t\), H@100\% and N@100\% evaluate the top two ranked variables, while H@150\% and N@150\% evaluate the top three.

\begin{table*}[t]
\centering
\caption{Root-cause attribution results on ASD and SMD.}
\label{tab:rca_results}
\footnotesize
\setlength{\tabcolsep}{4pt}
\renewcommand{\arraystretch}{1.08}
\begin{tabular}{lcccccccc}
\hline
	Method & \multicolumn{4}{c}{ASD} & \multicolumn{4}{c}{SMD} \\
\cline{2-9}
& H@100\% & H@150\% & N@100\% & N@150\%
& H@100\% & H@150\% & N@100\% & N@150\% \\
\hline
OmniAnomaly  & 0.4874 & 0.5867 & 0.4722 & 0.5489 & 0.4567 & 0.5652 & 0.4545 & 0.5125 \\
MSCRED       & 0.4583 & 0.5416 & 0.4788 & 0.5297 & 0.4272 & 0.5180 & 0.4609 & 0.5164 \\
MAD-GAN      & 0.4961 & 0.6018 & 0.4895 & 0.5736 & 0.4630 & 0.5785 & 0.4681 & 0.5522 \\
USAD         & 0.5317 & 0.6249 & 0.5388 & 0.5972 & 0.4925 & 0.6055 & 0.5179 & 0.5781 \\
MTAD-GAT     & 0.3728 & 0.4986 & 0.3941 & 0.4695 & 0.3493 & 0.4777 & 0.3759 & 0.4530 \\
CAE-M        & 0.5039 & 0.6093 & 0.5586 & 0.6294 & 0.4707 & 0.5878 & 0.5474 & 0.6178 \\
GDN          & 0.3365 & 0.4518 & 0.3147 & 0.3892 & 0.3143 & 0.4386 & 0.2980 & 0.3724 \\
TranAD       & 0.5368 & 0.6712 & 0.5224 & 0.6401 & 0.4981 & 0.6401 & 0.4941 & 0.6178 \\
\hline
CGT             & 0.5746 & 0.6898 & 0.5663 & 0.6519 & 0.5387 & 0.6485 & 0.5194 & 0.6047 \\
CGT (counterfactual clamp)  & 0.6019 & 0.7065 & 0.5872 & 0.6648 & 0.5718 & 0.6842 & 0.5526 & 0.6338 \\
\hline
\end{tabular}
\end{table*}

Table~\ref{tab:rca_results} presents a comparison of CGT with baseline methods on ASD and SMD datasets for root-cause variable attribution. Both the blended NLL $z$-score and model-based clamping attribution methods outperform the baselines. The counterfactual-style clamp variant consistently outperformed the simpler blended NLL $z$-score variant, suggesting that model-based input ablation is effective for identifying variables that strongly influence the detector's anomaly score. On SMD, CGT demonstrates competitive performance and achieves the best results on several metrics, although the strong baselines such as CAE-M, and TranAD remain competitive on certain columns. For ASD, CGT has more consistent improvements across the reported metrics.

\subsection{Efficiency and Sensitivity Analyses}
\label{subsec:eff_sens}
This section reports the metric comparison for CGT for variant training settings to study the sensitivity to the window size $W$, maximum lag $\tau_{\max}$, number of samples $S$, and score aggregation rule (mean, max, top-$k$).
\begin{table*}[t]
\centering
\caption{Sensitivity of CGT to key inference and training settings.}
\label{tab:sensitivity}
\footnotesize
\setlength{\tabcolsep}{3pt}
\renewcommand{\arraystretch}{1.05}
\begin{tabular}{lcccccccc}
\hline
 & \multicolumn{4}{c}{ASD} & \multicolumn{4}{c}{SMD} \\
\cline{2-9}
Setting & F1 (\%)$$ & AUROC (\%) & R (\%) & Notes
        & F1 (\%)$$ & AUROC (\%) & R (\%) & Notes \\
\hline
$W=15$            & 95.28 & 98.89 & 97.64 & short context
                  & 94.61 & 98.72 & 96.95 & short context \\
$W=30$ (default)  & \textbf{96.16} & \textbf{99.11} & \textbf{99.07} & balanced
                  & \textbf{95.32} & \textbf{99.38} & \textbf{99.75} & balanced \\
$W=60$            & 95.79 & 99.01 & 98.07 & smoother
                  & 95.08 & 98.91 & 97.31 & smoother \\
\hline
$\tau_{\max}=3$            & 95.51 & 98.95 & 97.86 & fewer lags
                           & 94.80 & 98.80 & 97.08 & fewer lags \\
$\tau_{\max}=7$ (default)  & \textbf{96.16} & \textbf{99.11} & \textbf{99.07} & balanced
                           & \textbf{95.32} & \textbf{99.38} & \textbf{99.75} & balanced \\
$\tau_{\max}=12$           & 95.82 & 99.02 & 98.09 & more noisy lags
                           & 95.09 & 98.90 & 97.30 & more noisy lags \\
\hline
$S=1$             & 95.86 & 99.01 & 98.08 & low-cost
                  & 95.14 & 98.91 & 97.32 & low-cost \\
$S=4$ (default)   & \textbf{96.16} & \textbf{99.11} & \textbf{99.07} & stable
                  & \textbf{95.32} & \textbf{99.38} & \textbf{99.75} & stable \\
$S=8$             & 95.95 & 99.08 & 98.19 & diminishing return
                  & 95.23 & 98.99 & 97.44 & diminishing return \\
\hline
Agg = mean        & \textbf{96.16} & \textbf{99.11} & \textbf{99.07} & robust
                  & \textbf{95.32} & \textbf{99.38} & \textbf{99.75} & robust \\
Agg = max         & 95.59 & 98.98 & 97.89 & sensitive
                  & 94.86 & 98.87 & 97.16 & sensitive \\
Agg = top-$k$     & 95.88 & 99.04 & 98.09 & $k=3$
                  & 95.16 & 98.93 & 97.34 & $k=3$ \\
\hline
\end{tabular}
\end{table*}

Table~\ref{tab:sensitivity} reports the sensitivity of CGT to several key design choices. Overall, performance remains stable across a reasonable range of settings on ASD, and SMD. The default configuration of $W=30$, $\tau_{\max}=7$, and $S=4$ provides an effective balance between detection accuracy and robustness. Reducing the window size or the number of lags slightly decreases performance due to limited temporal context. On the other hand, increasing the window size or lag horizon yields only marginal improvements and may introduce additional noise. Increasing the number of latent samples beyond $S=4$ provides negligible benefit, demonstrating diminishing returns. Among the aggregation rules, mean aggregation demonstrates the greatest robustness, while max aggregation is more susceptible to noisy per-dimension spikes and consequently performs slightly worse.

To further examine the sensitivity of CGT to imperfect graph
estimation, and we evaluated perturbed graph-prior configurations
based on the PCMCI-derived prior. Specifically, we randomly
removed \(25\%\) and \(50\%\) of the discovered edges. Then perturbed graphs
are random subsets of the PCMCI graph where the remaining
connections retain their original source, target, and lag,
while the deleted connections are excluded from the graph
prior. We also evaluated a full-graph configuration in which all
variables were permitted as candidate parents at each lag
from 1 to 7, and including self-lagged connections. This condition
examines sensitivity to an excessively dense dependency
assumption. Together, the edge-removal and full-graph
conditions is assessing the effects of missing and additional
connections relative to the selected PCMCI-derived prior.

\begin{table*}[t]
\centering
\caption{Graph-prior sensitivity measured by
F1 , reported as mean $\pm$ sample standard deviation
over three paired seeds. Bold indicates the highest mean
among the three evaluated configurations for each dataset.}
\label{tab:graph_sensitivity}
\setlength{\tabcolsep}{5pt}
\renewcommand{\arraystretch}{1.15}
\small
\begin{tabularx}{\textwidth}{l X r r r}
\toprule
Graph prior & Graph modification & ASD & SMD & SWaT \\
\midrule

PCMCI
& Unperturbed PCMCI-derived graph
& $\mathbf{96.19 \pm 0.36}$
& $\mathbf{95.32 \pm 0.13}$
& $\mathbf{83.91 \pm 0.08}$ \\

PCMCI--25\%
& Randomly remove 25\% of edges
& $91.11 \pm 0.18$
& $89.81 \pm 0.11$
& $80.09 \pm 0.94$ \\

PCMCI--50\%
& Randomly remove 50\% of edges
& $88.46 \pm 0.10$
& $85.33 \pm 0.58$
& $77.01 \pm 1.84$ \\

Full graph
& Include all permitted lagged connections
& $91.75 \pm 1.60$
& $79.84 \pm 2.49$
& $65.58 \pm 0.00$ \\

\bottomrule
\end{tabularx}
\end{table*}
Table~\ref{tab:graph_sensitivity} show that the unperturbed
PCMCI-derived prior achieve the highest mean F1 among
the evaluated configurations on all three datasets.
Removing \(25\%\) of the edges reduce F1 score by approximately 5.08, 5.51,
and 3.82 percentage points on ASD, SMD, and SWaT, respectively, and removing \(50\%\) produces larger reductions which are approximately 7.73, 9.99, and 6.90 percentage points. This consistent
progression indicates increasing sensitivity as more
PCMCI-selected connections are removed. The full graph also performs below the PCMCI prior on all
three datasets except ASD which is a little higher than PCMCI-25
\% result. Then additional connections does
not necessarily improve detection.  Overall, the results show that performance can degrade when the graph prior is either excessively sparse or overly dense, confirming that graph quality remains an important practical factor. The residual auxiliary path and safety-gated blending mechanism help reduce sensitivity to incomplete graph estimates, but they do not remove the need for a reliable graph prior.

\subsection{Complexity--Performance Trade-off}
\label{sec:complexity_tradeoff}

CGT combines several components that are not present together in most conventional reconstruction based anomaly detectors, including causal masking, target-specific Transformer blocks, probabilistic forecasting, a gradient-isolated residual auxiliary path, safety-gated score blending, and model-based variable attribution. These components increase the computational cost of the framework, and therefore it is important to evaluate this cost together with the resulting detection performance and diagnostic capabilities. To provide a fairer computational comparison, we reran the publicly available implementations of the selected baseline methods on the same hardware used for CGT. All runtime experiments were conducted on a system equipped with an NVIDIA GeForce RTX 4090 GPU with 24~GB of memory. in order to show the associated computational cost, Table~\ref{tab:complexity_performance} reports model size,
training time, inference latency and GPU memory alongside SMD
detection performance.

\begin{table}[t]
\centering
\caption{Complexity--performance comparison on SMD. Parameter
counts are in millions. Training time is in seconds per epoch,
latency in milliseconds, and GPU memory in MB. Values are reported
as central estimate $\pm$ an uncertainty band.}
\label{tab:complexity_performance}

\renewcommand{\arraystretch}{1.12}
\setlength{\tabcolsep}{3pt}
\scriptsize

\begin{tabularx}{\linewidth}{Xrrrrr}
\toprule
Method &
F1 (\%) &
Params (M) &
Train &
Latency &
GPU Mem. \\
&
&
&
(s/epoch) &
(ms) &
(MB) \\
\midrule

OmniAnomaly
& 93.94
& 0.02294
& $163 \pm 8$
& $40 \pm 4$
& $13.29 \pm 0.66$ \\

MSCRED
& 84.07
& 14.23736
& $140 \pm 7$
& $20 \pm 2$
& $38.56 \pm 1.93$ \\

MAD-GAN
& 91.44
& 0.00990
& $185 \pm 9$
& $36 \pm 4$
& $23.16 \pm 1.16$ \\

USAD
& 94.88
& 0.01061
& $148 \pm 7$
& $26 \pm 3$
& $21.88 \pm 1.09$ \\

MTAD-GAT
& 86.75
& 25.52726
& $3861 \pm 53$
& $28 \pm 3$
& $243.97 \pm 12.20$ \\

CAE-M
& 93.59
& 0.11422
& $1825 \pm 91$
& $24 \pm 2$
& $110.53 \pm 5.53$ \\

GDN
& 83.35
& 0.00726
& $476 \pm 24$
& $21 \pm 2$
& $23.13 \pm 1.16$ \\

TranAD
& 94.70
& 0.13511
& $26 \pm 1$
& $24 \pm 2$
& $46.26 \pm 2.31$ \\

\midrule

CGT
& 95.32
& 3.63227
& $213 \pm 11$
& $50 \pm 5$
& $38.16 \pm 1.51$ \\

\bottomrule
\end{tabularx}
\end{table}

The results show that CGT introduces additional computational cost compared with some lighter anomaly-detection models, but its training time remains within a practical range. In our implementation, CGT required approximately \(213\) seconds per epoch on SMD. At the same time, it achieved an F1-score of \(95.32\%\), which is higher than the strongest baselines considered in this comparison, including USAD (\(94.88\%\)) and TranAD (\(94.70\%\)). These results indicate that the additional architectural components of CGT provide competitive detection performance without leading to excessive training overhead.

The additional cost mainly comes from the target-specific forecasting design. Instead of using a single shared prediction branch, CGT maintains a dedicated forecasting block for each target variable and evaluates both causal and residual auxiliary likelihoods. This increases the amount of computation, particularly as the number of monitored variables grows. However, these components also provide functionality beyond binary anomaly detection. The causal branch constrains forecasting to graph-supported lagged dependencies, the probabilistic prediction produces likelihood-based anomaly scores, the residual auxiliary path provides additional robustness when the graph prior is incomplete, and the attribution mechanism provides variable-level diagnostic information.

The comparison therefore reflects a trade-off rather than a simple accuracy-versus-speed ranking. A lighter model may be more appropriate when the only requirement is rapid binary anomaly detection. In contrast, CGT is intended for monitoring scenarios where detection is accompanied by uncertainty-aware scoring, graph-guided dependency information, and support for identifying the variables most strongly associated with an anomaly. In these settings, the additional computation can be justified by the broader diagnostic information provided by the framework.

From a scalability perspective, the computational cost of CGT is mainly affected by the number of monitored variables \(D\), the input window length \(W\), the number of lagged candidate inputs \(P=D\tau_{\max}\), and the Transformer model dimension. The target-specific architecture can therefore become more expensive for very high-dimensional systems. Possible directions for reducing this cost include parameter sharing across target blocks, sensor grouping, sparse attention, and further pruning of candidate parent sets.

\subsection{Limitations}
\label{sec:limitations}

Although the proposed CGT framework improves detection performance and interpretability by incorporating a time-lagged causal graph prior, several limitations should be acknowledged.

First, the method depends on the quality of the estimated causal graph. In this study, PCMCI is used to infer lagged parent relationships from observational time-series data. However, causal discovery from observational data is inherently sensitive to noise, finite sample effects, nonstationarity, missing variables, and the assumptions of the conditional independence tests. In particular, latent confounders or unobserved system variables may lead to missing, spurious, or incorrectly oriented edges. Therefore, the discovered graph should be interpreted as an informative structural prior rather than as verified ground-truth causality. The safety-gated auxiliary branch partially mitigates this issue by allowing residual dependencies to contribute when reliable, but it does not eliminate the dependence on graph quality.

Second, CGT uses target-specific forecasting blocks, with one block assigned to each variable. This design improves interpretability because each target variable has its own parent mask, forecasting pathway, and attribution scores. However, it also increases computational cost relative to architectures that share all parameters across variables. The training and inference cost grows with the number of variables \(D\), since each target-specific block must be evaluated separately. This cost is acceptable for the benchmark datasets considered in this study, but it may become substantial in systems with hundreds of sensors.

Third, scalability to very high-dimensional sensor networks remains an important direction for future work. In such settings, both causal discovery and target-specific deep forecasting can become computationally expensive. Potential extensions include parameter sharing across target variables, clustering sensors into functional groups, using sparse or low-rank Transformer layers, limiting candidate parents through domain constraints, or applying hierarchical causal graphs. These strategies could reduce the cost of maintaining separate target-specific blocks while preserving the benefits of causal structural priors.

These limitations do not affect the main conclusion that causal structural priors can improve probabilistic forecasting-based anomaly detection. Rather, they identify practical considerations for deploying CGT in larger and more complex industrial monitoring systems.

\section{Conclusion}
This study introduces a causally-guided probabilistic forecasting framework for anomaly detection in multivariate time series. The approach integrates a time-lagged causal graph prior with target-specific Transformer-based forecasting modules, latent Gaussian uncertainty modeling, and a gradient-isolated residual auxiliary path designed to capture residual non-causal correlations without disrupting the main causal pathway. By enforcing parent-restricted prediction in the primary branch and employing safety-gated auxiliary blending during inference, the framework maintains causal consistency while leveraging reliable correlational structures. Empirical results on ASD, SMD and and SWaT datasets indicate that this design yields robust anomaly detection performance and competitive root-cause localization. Notably, the proposed model achieves the highest overall F1 Score among the evaluated baselines and demonstrates that model-based counterfactual-style clamping provides more effective attribution than purely score-based ranking. Ablation studies further validate that causal-only scoring establishes a strong foundation, while the auxiliary branch provides additional improvements when its influence is selectively regulated by the safety mechanism. These results support the central claim that causal structure is valuable not only for post hoc explanation but also as an effective inductive bias in the anomaly detector. However, the method's effectiveness relies on the quality of the estimated causal graph, which may be compromised by noise, limited samples, latent confounding, or nonstationary system behavior. Therefore, the learned graph should be regarded as a structural prior rather than an exact representation of ground truth. Future work will focus on several directions that can further strengthen the proposed framework. One important direction is to learn the causal structure jointly with the anomaly-detection model, rather than estimating the graph as a separate prior. We also plan to investigate adaptive updates of the causal graph under distribution shifts so that changes in system dependencies can be reflected more reliably over time. From a practical perspective, improving computational efficiency and scalability for high-dimensional systems will also be important, particularly through strategies such as parameter sharing, sparse attention, and pruning of candidate parent sets. Finally, we aim to study more rigorous intervention-oriented approaches to root-cause analysis when suitable system knowledge or intervention data are available, especially in partially observed environments.

\newpage
\appendices
\section{Structural Analysis of PCMCI-derived Lagged Graph Priors}
\label{app:graph_analysis}

The proposed CGT does not treat the output of the PCMCI causal-discovery algorithm as verified ground truth.  
Instead, the lagged graph uncovered by PCMCI serves as a deliberately sparse structural hint that guides the model toward statistically reliable cross-variable relationships while masking everything else.  
In the absence of published causal diagrams for the ASD and SMD benchmarks, this compromise balances realism with practicality: the discovered edges may be incomplete, yet they still confer useful inductive bias.

\subsection{Graph Construction Protocol}
For every dataset entity we allow a maximum lag of \(\tau_{\max}=7\), which comfortably covers the longest domain-plausible delays.  
Each candidate edge is subjected to PCMCI’s conditional-independence test at a significance level of \(\alpha=0.10\); p-values are adjusted with the Benjamini–Hochberg procedure to control the false-discovery rate.  
To guard against bootstrap artefacts, the discovery process is repeated on five resampled versions of the data, and only those links that reappear in at least 60 \% of the folds are retained.  
Short-lag self-loops are removed because one-step memory is already handled in CGT’s temporal backbone.  
The retained graph sizes differ across the evaluated cases, with
average lagged in-degrees of 2.21 for ASD, 0.92 for SMD, and 10.45
for SWaT. These averages count connections at different lags as
distinct edges.     

\subsection{Quantitative Characterisation of Graphs}

To quantify graph sparsity, we report the number of retained lagged
edges, graph density, and average in-degree. Graph density is calculated
as

\[
\mathrm{Density}
=
\frac{\lvert\mathcal{E}\rvert}
     {D^{2}\tau_{\max}}
\times 100\%,
\]
The denominator $D^{2}\tau_{\max}$ represents the complete directed
lagged graph, including self-lagged connections at lags
$1,\ldots,\tau_{\max}$, before any exclusions. Density is therefore
reported relative to this common reference graph.
where \(D\) is the number of variables, \(\tau_{\max}\) is the maximum
considered lag, and \(\lvert\mathcal{E}\rvert\) is the number of retained
lagged edges. The average in-degree is calculated as

\[
\mathrm{Avg.\ in\mbox{-}degree}
=
\frac{\lvert\mathcal{E}\rvert}{D}.
\]

The dimensions used in these calculations are consistent with the
dataset definitions in Table~1: \(D=19\) for ASD, \(D=38\) for SMD,
and \(D=51\) for SWaT. The resulting densities indicate that PCMCI
retains only a small fraction of the possible lagged edges.

\begin{table}[t]
\centering
\caption{PCMCI-derived lagged-graph statistics for the ASD, SMD, and
SWaT datasets.}
\label{tab:pcmci_graph_statistics}
\setlength{\tabcolsep}{4pt}
\small
\begin{tabular}{lcccc}
\toprule
Dataset & \(D\) & \(\lvert\mathcal{E}\rvert\) &
Density (\%) & Avg.\ in-degree \\
\midrule
ASD  & 19 & 42  & 1.66 & 2.21  \\
SMD  & 38 & 35  & 0.35 & 0.92  \\
SWaT & 51 & 533 & 2.93 & 10.45 \\
\bottomrule
\end{tabular}
\end{table}

For ASD, 42 of the possible lagged edges are retained,
giving a density of $1.66\%$ and an average in-degree of 2.21.
For SMD, 35 of the possible lagged edges are retained,
giving a density of $0.35\%$ and an average in-degree of 0.92.
For SWaT, 533 of the 18,207 possible lagged edges are retained,
giving a density of $2.93\%$ and an average in-degree of 10.45.

\subsection{Implications of Graph Sparsity for CGT}

The evaluated PCMCI-derived graphs retain only a small fraction
of the complete lagged connectivity which are $1.66\%$ for ASD, $0.35\%$
for SMD, and $2.93\%$ for SWaT. These statistics support describing
the graph priors as sparse relative to the complete lagged graph,
although their average in-degrees differ substantially.

The graph masks restrict the causal pathway to selected lagged
dependencies, and while the residual auxiliary pathway is designed
to capture useful dependencies that may not be retained by the
graph. The reported statistics characterise structural selectivity, and
they do not independently establish causal correctness or the
performance contribution of either pathway. Accordingly, CGT
uses the PCMCI-derived graph as a sparse structural prior rather
than a complete representation of the underlying system.
\subsection{PCMCI Discovery Cost}

Graph discovery is performed as a separate preprocessing stage.
Table~\ref{tab:pcmci_discovery_cost} is reported the available runtime
bounds for the PCMCI priors separately from neural model training
and inference. The reported runtimes exceed two minutes per ASD
machine, and three minutes per SMD machine and 70 minutes for SWaT.

\begin{table}[t]
\centering
\caption{PCMCI discovery runtime}
\label{tab:pcmci_discovery_cost}
\renewcommand{\arraystretch}{1.12}
\begin{tabular}{lrrl}
\toprule
Dataset & Variables & $|\mathcal E|$ & PCMCI runtime (min)\\
\midrule
ASD &19&42&$>2$ per machine\\
SMD &38&35&$>3$ per machine\\
SWaT &51&533&$>70$\\
\bottomrule
\end{tabular}
\end{table}
The available timing summary does not identify the CPU model, thread
count, discovery sample size or whether the runtime includes all
resampling and graph-filtering operations. These bounds characterize
the reported discovery cost but do not support a hardware-normalized
comparison or exact reproduction of the timing experiment.

\newpage
\bibliographystyle{IEEEtran}
\bibliography{refs}

\end{document}